# BUILDING ONTOLOGIES TO UNDERSTAND SPOKEN TUNISIAN DIALECT

Marwa GRAJA[1] and Maher Jaoua[1] and Lamia Hadrich Belguith[1]

[1]Miracl Laboratory, ANLP Research Group, University of Sfax, Tunisia.
http://sites.google.com/site/anlprg
www.miracl.rnu.tn
marwa.graja@fsegs.rnu.tn, maher.jaoua@fsegs.rnu.tn,
l.belguith@fsegs.rnu.tn

## ABSTRACT

*This paper presents a method to understand spoken Tunisian dialect based on lexical semantic. This method takes into account the specificity of the Tunisian dialect which has no linguistic processing tools. This method is ontology-based which allows exploiting the ontological concepts for semantic annotation and ontological relations for speech interpretation. This combination increases the rate of comprehension and limits the dependence on linguistic resources. This paper also details the process of building the ontology used for annotation and interpretation of Tunisian dialect in the context of speech understanding in dialogue systems for restricted domain.*

## 1. INTRODUCTION

The semantic parsing of understanding component in the context of dialogue systems helps to clarify the utterances meaning [1]. Indeed, the understanding component should provide a semantic interpretation of the meaning, taking into account the specificity of the spontaneous speech and the spoken dialect.

Many approaches have been proposed to understand utterances in spoken dialogue. Among them, we mention syntactic approach [2] and semantic approach [3][4]. All of theses approaches could be based on stochastic methods or on rules and patterns methods.

Since the spoken dialects are not written, it is very hard to obtain adequate corpora to deal with stochastic methods. In fact, one of the major limitations of the stochastic methods is that it relies on a large quantity of annotated data [5]. Moreover, tools of speech transcription to obtain transcripted corpora are very expensive especially for Arabic dialect. On the other hand, using methods based on rules and parsers to semantic parsing could pose inefficiency problem [6] especially in the case of a restricted domain. In fact, users often use keywords rather than well-structured sentences [7]. These characteristics are identified in a spoken dialogue for many restricted domain. Indeed, we noted the extensive use of keywords and the neglect of grammatical structures. Another important problem is the lack of resources for the studied dialect. In fact, there are no resources available for the Tunisian dialect. These observations led us to think about a lexical semantic approach to build the understanding component for the Tunisian dialect. The method described in this work explores the meanings of words and their interconnection based on lexical choice databases. This is the subject of lexical semantics researches [8]. In this work, the lexical choice consists of integrating the ontology as a knowledge base which allows annotation and semantic interpretation of utterances. Indeed, Ontologies have been used in several systems for semantic annotation. In this context, Allen and al. [9] use a domain independent ontology to provide semantic representation in understanding module of a dialogue system. However, this work is mainly based on grammars to identify linguistic relations. We also mention the work of [7] which uses Ontologies for the generation of a command in natural language. The work of Milward [10] also uses Ontologies to increase the level of clarification in dialogue manager. In fact, Milward affirms that the use of Ontologies in dialogue systems could reduce the dependence on hand-crafted linguistic components.

Our contribution in this work consists in processing Tunisian dialect in dialogue systems which has no resources. To our knowledge, this work is the first which deals with Tunisian dialect in an understanding module of a dialogue system. In this work we use domain Ontology to cover the lexicon used in the all services of railway station and we use a task Ontology which gathers all achievable tasks in this area as request information about the train time, booking, etc. To process utterances in Tunisian dialect in the restricted field, we exploit Ontologies for the semantic annotation and interpretation. Semantic annotation is used to assign semantic labels to each word without making a relationship between words or a group of words, while the semantic interpretation has a purpose larger than the semantic annotation. Indeed, the semantic interpretation increases the rate of comprehension by emphasizing the relationships between words in the same utterance.

This paper is organized as follows. The next two sections present an overview of Tunisian dialect and Ontologies. Section 4 presents the methodology used for building our Ontologies. Sections 5 and 6 present our method for the annotation and semantic interpretation based on Ontologies. Sections 7 and 8 show the implementation and report results of this work. Finally, conclusion is presented in section 9.

## 2. TUNISIAN DIALECT

The Arabic language is a collection of several variants including Modern Standard Arabic (MSA) which has a special status as an official standard of the Arab world. It is the main language of media and culture. Other variants of Arabic are dialects which are spoken and informal.
The automatic processing of a dialect is a big challenge, because it has some resemblance to the MSA and adds a great variety. Indeed, Arabic people consider the dialect as a variety of standard Arabic. But from a linguistic point of view, it can be seen as a full standard [11]. This justifies the classification of dialect in another language. So, the tools developed for standard Arabic can not be used in the dialect given the major difference between standard Arabic and dialect in terms of phonology, morphology, syntax and lexical level. Consequently, we could not use standard Arabic tools and adapt them to parse Arabic dialect.

The Arabic dialect represents the real form of language. They are generally limited in use for informal everyday communication [11]. In fact, the dialect is mainly spoken and unwritten. So, it is crucial to study Arabic dialects in dialogue systems because there are few studies which deal with and it is a standard in spoken dialogue. We have chosen the Tunisian dialect as a representative example to study Arabic dialects in dialogue systems. Also, it should be noted that this work investigates the Tunisian dialect assuming the absence of all resources and tools for Tunisian dialect processing.

The Tunisian dialect is characterized by many features especially in a restricted field. To begin with, utterances in Tunisian dialect are not very long. In fact, the average of word number in an utterance is about 3.61 [12]. Another feature in Tunisian dialect is the non respect of grammar and the use of foreign words especially in the studied field. In fact the most important key words are borrowings from French language. Finally, user utterances are characterized by a frequent use of domain dependant key words. All these features led us to investigate a method which focus on key word and relation between them and use a knowledge base to annotate and interpret utterances in Tunisian dialect by means of Ontologies.

## 3. ONTOLOGIES

Ontology is a formal specification, explicit and consensual conceptualization of a domain [13]. Indeed, the design and creation of Ontologies help humans to understand and solve the ambiguities for specific domains [14]. It consists of a set of concepts linked together in a methodological manner. This relationship can be made using the taxonomic relationships (subclass) or non-taxonomic relations. Non-taxonomic relations are semantic relations that can be added to describe the special relationship in a well determined domain.

In the literature, we can identify several types of Ontologies. In this work, we are interested in specialized Ontologies which are domain ontology and task ontology. Indeed, the domain ontology provides the vocabulary of concepts and terminology of the domain [10] and the interactions between these concepts in the concerned field. These are reusable Ontologies within a given domain, but not from one domain to another. While task ontology contains all tasks performed in a given domain [16]. According to [17], the task ontology describes a vocabulary related to a task.

It should be noted that the use of existing Ontologies is crucial. Nevertheless, we have no domain ontology in Tunisian dialect in the studied field. Therefore, we have manually built ontology by following a known methodology.

## 4. ONTOLOGIES BUILDING

Several methodologies for building ontology have been identified [17][18]. We can mention as an example MethoOntology methodology and OntoClean methodology. There are other methodologies which are proposed for the construction of linguistic ontology. TERMINAE methodology is one of the methodologies which allows manual construction of Ontologies from texts, based on language processing tools, to extract the lexicon and lexical or syntax relations. However, these tools deal only with French and English languages. So we can not use tools of this methodology, but we will only use the general approach because it meets to our needs.
The main purpose in this section is to explain the various steps followed in this work to build Ontologies using the TERMINAE methodology. The designed Ontologies will be integrated into the understanding component for semantic interpretation of utterances in the context of a spoken dialogue system.

To start the construction of our Ontology, we should have a corpus covering the treatment area and representing the domain knowledge. The used corpus is called TuDiCoI (Tunisian Dialect Corpus interlocutor) which is a corpus of spoken dialogue in Tunisian dialect and it is obtained from a railway information service. It is a pilot corpus which gathers a set of conversations recorded in the railway station between the staff and customers who request information about the departure time, price, booking, etc [12]. It is a corpus which represents the lexicon used in conversations in the field of railway information.

In an understanding module of a dialogue system, we are interested in user utterances. That's why; we process in this corpus only user utterances. So, among 369 as user utterances, we considered randomly 194 utterances as a development corpus which consists of 701 words. The rest of the utterances are left as a test corpus. We have manually transcribed the corpus because the lack of resources for automatic dialogue transcription especially for Tunisian dialect.

Table 1. Development corpus statistics

| Utterance (U) | Word (W) | Average (W/U) |
|---------------|----------|---------------|
| 194 | 701 | 3.61 |

After fixing the goal of the ontology and the domain corpus; we will continue the TERMINAE process to manually build the ontology from a transcribed spoken corpus.

## 4.1. Lexical specification

The lexical specification step consists of extracting the representative lexical of the domain and relations between lexical varieties. It is based on linguistic analysis of the corpus. Since we do not have language tools for Tunisian dialect analysis, we have to do manual linguistic analysis to extract the lexicon and lexical relations. This step requires the intervention of linguistic experts to validate the obtained lexicon and lexical relations. After lexicon specification, we try to classify by group each set of semantic lexicon carrying the same semantic. For example, { نورمال (1), أسريع (2), لكسبراس (3)} {nwrmAl (1), Os~ryE (2), lksbrAs (3)}[1] {Normal (1), Rapid (2), Express (3)} is a group which is carrying the same semantic, it is the lexicon used to specify the train type. the lexical group {يخرج (1), دييار (2), يمشي (3)} {leaves (1), departure (2), parting (3)} {yxrj (1), dyPAr (2), ym$y (3)} is another example which represents the lexical links between the train and words used to express a departure time.

## 4.2. Standardization

This step consists of assigning to each lexical variation a concept. So, from lexical varieties and lexical relations, we obtained a group of concepts and semantic relationships. At the end of this phase, we get a semantic network represented by a set of concepts linked by semantic relations. As example of standardization, the lexical group {نورمال (1), أسريع (2), لكسبراس (3)} {nwrmAl (1), Os~ryE (2), lksbrAs (3)} {Normal (1), Rapid (2), Express (3)} is denoted by the concept "Train_Type". The standardazition step is only applied to each group of lexical variation and not to lexical relationships. So, we consider all relationship.

We noticed through the linguistic study of our corpus that every departure city is preceded by the linguistic mark "من" "*mn*" "*from*" to identify a departure city. It is the only lexical group for this case. So, we consider this marker as a semantic relation between the concepts "Train" and "Departure_City".

---

[1] For all examples, the transliteration is produced by the Buckwalter Arabic Transliteration System (http://www.qamus.org/transliteration.htm).

About the number of concepts in both Ontologies, we have identified 6 concepts in the task Ontologies which are "Path_Request", "Departure_Time_Request", "Arrival_Time_Request", "Availability _Request", "Price_Request", and "Bookin_Request". The first 3 concepts concern the "Train" concept and the others concern the "Ticket" concept. About the domain ontology, we have identified 15 concepts which are "Train", "Departure_City", "Arrival_City", "Departure_Hour", "Arrival_Hour", "Train_Type", "Ticket", "Ticket_Class", "Ticket_Type", "Ticket_Price", "Ticket_Number". Under the "Arrival_Hour" and "Departure_Hour" concepts, we have identified 4 sub concepts which are "Exact_Day", "Relative_Day", "Exact_Hour", and "Relative_Hour".

### 4.3. Formalization

The purpose of the formalization step is to translate the semantic network obtained in the previous step into a knowledge representation language. In our work, the formalization is done by the OWL language (Ontology Web Language)[2]. OWL is the standard currently proposed by the W3C for representing Ontologies. Indeed, OWL constitutes a knowledge representation language used to represent knowledge in a form usable by the machine (i.e. make the ontology accessible and understandable by the machine). OWL has many advantages. Among them, it has richer expressivity than other languages such as XML and RDF.

## 5. SEMANTIC ANNOTATION

The semantic annotation is defined as the process used to associate semantic labels to each word or a group of words in a statement. In this work, we perform a semantic annotation of a transcribed speech based on domain ontology and task ontology. These two Ontologies are the sources of knowledge for semantic annotation. Hence, it is important to note that there are many problems in the utterances annotation since the dialect is the target of this study. Indeed, the lexical variety of spoken dialect between standard Arabic, foreign words, dialect words and disfluencies and absence of the dictionary are disabilities for the treatment. Also, we note that the spoken dialect does not respect correct grammatical forms, which prevents us to use any analyzer of standard Arabic language and adapt it to the Tunisian dialect. The second problem is the segmentation of the utterance that appears as a key critic step in semantic annotation. Indeed, it sometimes becomes difficult to identify the text elements to annotate [15] because of morphological varieties of words, compound words, etc. So we have to make a standardization step of statements before doing ontology-based annotation by following the same standard used to build the ontology in order to make a correspondence between elements of the ontology with words of the utterance to be labeled.

- Treatment of compound words

Through the corpus study, we identified a definite list of compound words in the domain of railway information. This list is kept in a compound words dictionary to facilitate its detection during the analysis. We have identified 55 significant compound words of the studied domain

- Radicalization and lexical variants removal

The radicalization step to remove lexical variants [18] is an effective method especially in the case of a limited field. Indeed, we try to deduce the singular form if the word is in the plural and use the base form in the case of morphological variants.

---

After utterance normalization, we attribute a semantic label for each word in the utterance based on the domain ontology and the task ontology. In fact, we exploit our ontologies by scanning all concepts instances of both Ontologies and we look for the presence of a given word in the ontology instances. It should be noted that it is possible to have two concepts, or one concept or no concept attributed to a given word. In case of having two different labels for a given word, it is possible to improve understanding through the phase of semantic interpretation which is presented in the next section. In case of a single semantic label for a word, we noticed a percentage of 95% that the attributed label to a given word is correct. The case of non recognition of the semantic label of a given word could be explained by incomplete words, a non-domain lexicon and peculiar phenomena to spontaneous speech which are not yet studied in this work. It can be noted in the example of figure 1 that the first word has two different semantic labels. So we should ameliorate semantic annotation by a semantic interpretation step to disambiguate the meaning.

The use of ontology in the semantic annotation step does not provide significant benefit and its use in this step does not exceed the use of a domain dictionary. But the major contribution of the use of Ontologies is at the semantic interpretation presented in the next section. Therefore, the next step is a semantic interpretation which improves the semantic annotation by using the semantic relations of two Ontologies.

```
                                            وقتاش يخرج القطران؟
                                         wqtAš yxrj OtrAn
                                    When does the train leave?
    <token value='وقتاش '>
    <annotation>Departure_Time_Request </annotation>
    <annotation>Arrival_Time_Request </annotation>
    </token>
    <token value='يخرج '>
    <annotation>Semantic_Relation</annotation>
    </token>
    <token value='القطران '>
    <annotation>Train</annotation>
    </token>
```

Figure 1. Example of semantic annotation

## 6. SEMANTIC INTERPRETATION

The semantic interpretation can be defined as a semantic decoding which clarifies the semantics carried in a statement and increases the accuracy of understanding.

It is important to note that this work is not interested in contextual interpretation; it is only interested in lateral interpretation (i.e. context-free dialogue). Indeed, we try to improve label utterance's words taking into account only semantic relations between the words of the same utterance and not an interpretation based on all utterances in the same dialogue. In this step, we try to improve the semantic annotation of the first step by exploiting the semantic relationships between words in the same statement. Indeed, the semantic interpretation is done to semantically link words together and identify semantic relations in the utterance. These semantic relations should have correspondence in the ontology to help expressing a precise semantics.

The semantic interpretation is triggered for each word with two different labels detected after the semantic annotation step. Indeed, when we detect two labels for the same word, we traverse the utterance to detect the semantic relations in the utterance. If a relationship is identified in the utterance, we check in the ontology if the target of this relationship is one of the concepts already identified as a label tag for a word or not. If yes, the target of the semantic relation is the correct label and should be attributed as a label to the word.

To explain the proposed method for semantic interpretation, we take the following utterance as an example: "ساعه إلماضي تونس لتونس" "ltwns IlmADy sAEh" "To Tunis One Hour o'clock". First, we begin by the standardization step which consists of detecting the compound words and the semantic relationships. The statement becomes: "ساعه_إلماضي تونس إلى" "IlY twns IlmADy_sAEh" "To Tunis One Hour o'clock". After the standardization step, we annotate semantically an utterance as it is presented in figure 2.

```
<token value='إل '>
<annotation>Semantic_Relation</annotation>
</token>
<token value=' تونس '>
<annotation>Arrival_City</annotation>
<annotation>Departure_City</annotation>
</token>
<token value=' الماضي_ساعه '>
<annotation>Hour</annotation>
</token>
```

Figure 2. Example of semantic annotation of the utterance "ساعه_إلماضي تونس إلى"

We note in figure 2 that the word "تونس" "twns" "Tunis" has two different semantic labels. On the one hand, it is labelled as "Departure_City" and on the other hand it is labelled as "Arrival_City". Now, we apply the semantic interpretation step. To explain this point, we present an extract of our domain ontology in a semantic network representation illustrated in figure 3. In this figure, the instance "تونس" "Tunis" belongs to two different concepts. At the interpretation level, we use semantic relations. In the utterance, there is the relationship "إلى" "to" which its target is the concept "الوصول مدينة" "Arrival_City". So the word "تونس" "Tunis" should have as semantic label "الوصول _ مدينة" "Arrival_City".

Moreover, semantic relation can be more useful for critical semantic interpretation. We explain this trough the following example "أن كلاس" "class one". The word "أن" "Un" "One" is annotated as "Ticket_Number" and "Ticket_Class" at the same time. In this step, we focus on semantic relation to clarify the meaning of the word. In fact, we have a semantic relation "كلاس" "Class" which can be useful in such situation. So, the word "أن" is finally annotated as "Ticket_Class" because it is the target of the semantic relation "كلاس" "Class" in the domain ontology.

During the detection of semantic relations in the utterance, it is possible to find more than one relation. So we followed a strategy in the development which consists of taking into account the closest relations from the right or the left of the word to be labelled. Once a relationship is used in the interpretation, it will be ignored in the following interpretations.

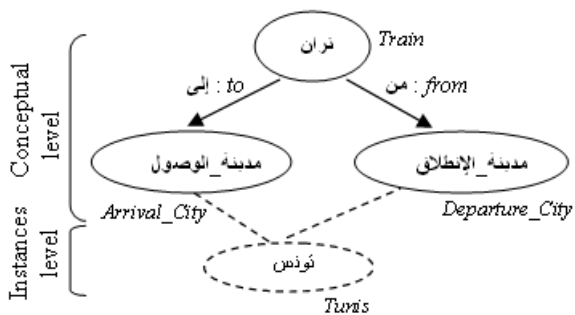

Figure 3. Extract from the domain ontology in a semantic network representation

## 7. IMPLEMENTATION

To build our Ontologies, we used the Protégé platform. It is an open-source platform that provides a growing user community with a suite of tools to build domain models and knowledge-based applications with ontology[3].

It is a modular interface, developed at Stanford Medical Informatics at Stanford University, for editing, viewing, checking (supervision constraints) Ontologies [19]. The knowledge model of Protégé contains classes (concepts), slots (properties) and facets (property values and constraints), as well as instances of classes and properties. Many plug-in are available or can be created by the user. Among these, we can cite the plug-in which can be used to OWL language and other plug-ins for ontology visualization. Protégé allows the generations of OWL file which represent the ontology. In fact, OWL language is used in our work as a formalization language. Indeed, in an OWL language each concept is represented by an OWL class `<owl:Class rdf:about= "label">`. The label is a unique identifier of a concept. The concepts are linked together by taxonomic relations which are represented in OWL by `<rdfs:subClassOf>`. The concepts can also be connected by non-taxonomic relations which are specific relations to a given field. These relations are represented by the property `<owl:objectProperty>`. This type of relation should have an original concept and an arrival concept represented in an OWL language respectively by `<rdfs:domain>` and `<rdfs:range>`. To implement our method of semantic interpretation, we used the Jena framework. It is a Java framework for building Semantic Web applications. It provides a programmatic environment for RDF, RDFS and OWL and includes a rule-based inference engine. Jena is open source and grown out of work with the HP Labs Semantic Web Programme[4]. The Jena Framework includes an OWL API[5] which facilitates the operating of the ontology.

## 8. RESULTS

Our test corpus consists of 175 user utterances. This corpus is manually transcribed in the same way as the development corpus. This corpus contains 670 words.

To evaluate the results of our method of semantic interpretation, we used F-measure and Precision. These measures are used to measure the semantic labels assigned to different words in oral utterances with relation to instances of ontology concepts. In our case, the Precision measures the number of words correctly labelled divided by the total number of annotated words (correctly labelled and not correctly labelled) and F-measure measures the number of words correctly labelled divided by the total number of words of the test corpus.

Note that the number of words correctly recognized is 448, and the number of words which are not recognized is 208 words. So we get a Precision of 0.96 and an F-measure of 0.66

Table 2. Experimentation results on the test corpus

---

| Correct Annotation | *(a)* | 448 |
|---|---|---|
| Incorrect Annotation | *(b)* | 14 |
| Not Recognized | *(c)* | 208 |
| Total | *(d)* | 670 |
| **F-Measure** | *(a / d)* | 0.66 |
| **Precision** | *(a/(a+b))* | 0.96 |

The precision ratio obtained for this evaluation is encouraging because we have not yet dealt with the specific phenomena of spontaneous speech. Indeed, after the analysis of 208 tokens which are not recognized in the interpretation, we noted that 38 tokens come from speech phenomena such as hesitation, incomplete words and the rest represent anaphors and out of vocabulary lexicon.

## 9. CONCLUSION

In this work, we have proposed a method which takes into account the specificity of the Tunisian dialect which has no linguistic processing resources. Indeed, the proposed method is based on lexical semantics which incorporates domain ontology and task ontology for the semantic interpretation of the utterance in a spoken dialogue without incorporating methods based on rules or parsers. In this method, we used the concepts of both Ontologies to annotate the utterances, while the semantic relations of Ontologies are used to disambiguate the interpretation and to increase the understanding level. To our knowledge, this method is the pioneer which proposes understanding the Tunisian dialect in a limited domain. The proposed method is implemented and tested on a Tunisian dialect corpus using specialized framework in the Ontologies processing. Results are encouraging for a first validation of this method. Indeed, we obtained an accuracy of 0.96. This precision is being improved by incorporating a processing level of particular phenomena of spontaneous speech.

## REFERENCES

[1]  Minker W. (1998) "Stochastic versus rule-based speech understanding for information retrieval". Speech Communication, pp 223-247.

[2]  Goulian J., Antoine J.Y, (2001) "Compréhension Automatique de la Parole combinant syntaxe locale et sémantique globale pour une CHM portant sur des tâches relativement complexes", TALN, Tours, 2-5 July.

[3]  Hadrich Belguith  L., Bahou  Y., Ben Hamadou  A.B. (2009)  "Une méthode  guidée par la sémantique pour la compréhension Automatique des énonces oraux arabes",  International journal of information sciences for decision making (ISDM), N  35, September.

[4]  Wong Y.W and  Mooney R.J. (2006)  "Learning for Semantic Parsing with Statistical Machine Translation", in HLT-NAACL '06 Proceedings of the main conference on Human Language Technology Conference of the North American Chapter of the Association of Computational Linguistics.

[5]  Wong Y. W. (2005) "Learning for semantic parsing using statistical machine translation techniques". Technical Report UT-AI-05-323, Artificial Intelligence Lab, University of Texas at Austin, Austin, TX, October.

[6]  Sabouret N. and Mazuel L, (2005) "Commande en langage naturel d'agents VDL". In Proceedings of WACA'05, p. 53–62.

[7]  Mazuel L., (2007) "Utilisation des ontologies pour la modélisation logique d'une commande en langue naturelle". ColloRECITAL, Toulouse.

[8]  Boite R. and  Kunt M., (1987)  "Traitement de la parole",  Presses Polytechniques Romandes, Paris, France.

[9]    Allen J., Dzikovska M., Manshadi M., and Swift M., (2007) "Deep linguistic processing for spoken dialogue system". In proceedings of the Workshop on Deep Linguistic Processing (DeepLP '07).

[10   ]Milward D. and Beveridge M., (2003) "Ontology-based dialogue systems". In proceedings of the 3rd workshop on knowledge and reasoning in practical dialogue systems (IJCAI 03), Mexico.

[11]  Diab M. and Habash N., (2007) "Arabic Dialect Processing Tutorial". Proceedings of the Human Language Technology Conference of the North American. Rochester.

[12]  Graja M., Jaoua M. and Hadrich Belguith L., (2010) "Lexical Study of A Spoken Dialogue Corpus in Tunisian Dialect". The International Arab Conference on Information Technology (ACIT), Benghazi – Libya.

[13]  Gruber T.R., (1993) "A translation approach to portable ontology specifications", Knowledge Acquisition, 5 (2), pp 199-220.

[14]  Ghorbel H., Bahri A., and Bouaziz R., (2008) "Les langages de description des ontologies: RDF & OWL". Acte des huitièmes journées scientifique des jeunes chercheurs en Génie Electrique et Informatique (GEI), Sousse-Tunisia.

[15]  Ma Y., Audibert L. and Nazarenko A., (2009) "Ontologies pour l'annotation sémantique", Ingénieries des connaissances.

[16]  Mizoguchi R.,Vanwelkenhuysen J. and Ikeda M., (1995) "Task Ontology for Reuse of Problem Solving Knowledge". In proceedings of the 2nd International Conference on Building and Sharing of Very Large-Scale Knowledge Bases. (KB & KS'95).

[17]  Guarino N., (1998) "Formal Ontologies and information systems". In proceedings of FOIS'98, IOS Press, Amsterdam

[18]  Rijsbergen C.J.V. (1979) "Information Retrieval. Butterworths", London

[19]  Noy N., Ferguson R.W. and Musen M.A.(2000) "The Knowledge Model of Protégé-2000: Combining Interoperability and Flexibility". In Proceedings of the 12th European Knowledge Acquisition Workshop (EKAW'00).

## Authors

**Marwa GRAJA:** received her Engineer degree in Computer Science engineering from the National School of Engineers of Sfax-Tunisia (ENIS) in 2006, and Master degree in Computer Science (NTSID) from the National Engineering School of Sfax-Tunisia (ENIS) in 2008. Actually, she is persuading her PhD studies with ANLP Research Group within MIRACL Laboratory. Her research interests include Ontology, Tunisian Dialect, and Dialogue systems.

**Maher JAOUA:** received his Engineer degree and master degree in Computer Science at the University of Tunis-Tunisia in respectively 1995 and 1997. He received his Ph.D. degree in Computer Science from the University of Tunis in 2004. He is currently an assistant professor of Computer Science at the Faculty of Economic Sciences and Management of Sfax (FSEGS) - University of Sfax (Tunisia).  He is a member of ANLP (Arabic Natural Language Processing) Research Group within MIRACL Laboratory since 1997. His main research activities are focused on text summarization and dialogue systems.

**Lamia HADRICH BELGUITH:** received her postgraduate diploma in Computer Science at Faculty of Economics and Management of Sfax (FSEGS) in 1990. She received her master degree in Management Information Systems at High School of Management -Tunis (ISG) in 1992. Then, she received PhD degree from the Faculty of Sciences -Tunis, Tunisia, in 1999. From 1999 to 2009, she was an Associate Professor of Computer Science and Management at Faculty of Economic science and Management (FSEGS), University of Sfax. Since 2009, she is a professor of Computer Science and Management at FSEGS and head of Arabic Natural language Research Group (ANLP-RG) of Multimedia, InfoRmation systems and Advanced Computing Laboratory (MIRACL). Her research activities have been devoted to several topics: Arabic text analysis, Automatic Abstracting, Question-Answering systems, and Human-machine spoken dialogue systems.